\documentclass[conference]{IEEEtran}
\IEEEoverridecommandlockouts
\usepackage{cite}
\usepackage{amsmath, amssymb, amsfonts}
\usepackage{graphicx}
\usepackage{textcomp}
\usepackage{array}
\usepackage{booktabs}
\usepackage{xcolor}
\usepackage{enumitem}
\usepackage{tabularx}
\let\oldthebibliography\thebibliography
\renewcommand\thebibliography[1]{
  \oldthebibliography{#1}
  \setlength{\itemsep}{1mm}
}
\usepackage{hyperref}
\hypersetup{
    colorlinks=true,
    linkcolor=black,
    citecolor=black,
    urlcolor=black
}
\usepackage{url}
\usepackage[a4paper, margin=2.54cm, right=1.91cm]{geometry} 

\usepackage[ruled]{algorithm2e}
\def\BibTeX{{\rm B\kern-.05em{\sc i\kern-.025em b}\kern-.08em
    T\kern-.1667em\lower.7ex\hbox{E}\kern-.125emX}}
\begin{document}

\makeatletter
\newcommand{\linebreakand}{%
  \end{@IEEEauthorhalign}
  \hfill\mbox{}\par
  \mbox{}\hfill\begin{@IEEEauthorhalign}
}
\makeatother

\title{Differential Evolution Algorithm based Hyper-Parameters Selection of Transformer Neural Network Model for Load Forecasting}

\author{
\IEEEauthorblockN{1\textsuperscript{st} Anuvab Sen}
\IEEEauthorblockA{\textit{Electronics and Telecommunication} \\
\textit{Indian Institute of Engineering}\\
\textit{Science and Technology, Shibpur}\\
Howrah, India\\
sen.anuvab@gmail.com}
\and
\IEEEauthorblockN{2\textsuperscript{rd} Arul Rhik Mazumder}
\textit{Computer Science} \\
\IEEEauthorblockA{\textit{School of Computer Science} \\
\textit{Carnegie Mellon University}\\
Pittsburgh, United States of America \\
arul.rhik@gmail.com}
\and
\IEEEauthorblockN{3\textsuperscript{rd} Udayon Sen}
\IEEEauthorblockA{\textit{ Computer Science and Technology} \\
\textit{Indian Institute of Engineering}\\
\textit{Science and Technology, Shibpur}\\
Howrah, India\\
udayon.sen@gmail.com} 
}

\IEEEoverridecommandlockouts
\IEEEpubid{\makebox[\columnwidth]{Copyright: 978-1-5386-5541-2/18/\$31.00~\copyright2023 IEEE \hfill} \hspace{\columnsep}\makebox[\columnwidth]{ }}

\maketitle

\IEEEpubidadjcol

\begin{abstract}
Accurate load forecasting plays a vital role in numerous sectors, but accurately capturing the complex dynamics of dynamic power systems remains a challenge for traditional statistical models. For these reasons, time-series models (ARIMA) and deep-learning models (ANN, LSTM, GRU, etc.) are commonly deployed and often experience higher success. In this paper, we analyze the efficacy of the recently developed Transformer-based Neural Network model in load forecasting. Transformer models have the potential to improve load forecasting because of their ability to learn long-range dependencies derived from their Attention Mechanism. We apply several metaheuristics namely Differential Evolution to find the optimal hyperparameters of the Transformer-based Neural Network to produce accurate forecasts. Differential Evolution provides scalable, robust, global solutions to non-differentiable, multi-objective, or constrained optimization problems. Our work compares the proposed Transformer-based Neural Network model integrated with different metaheuristic algorithms by their performance in load forecasting based on numerical metrics such as Mean Squared Error (MSE) and Mean Absolute Percentage Error (MAPE). Our findings demonstrate the potential of metaheuristic-enhanced Transformer-based Neural Network models in load forecasting accuracy and provide optimal hyperparameters for each model.
\end{abstract}


\section{Introduction}
Load forecasting is the application of science and technology to predict the future demand for electricity or power in a given geographical location, for some specific future time. It plays a crucial role in various sectors, such as energy, trading and markets, infrastructure planning, disaster management, etc., to name a few. Traditional load prediction methods rely on historical data and models that simulate patterns of electricity consumption, but such models often face challenges in accurately capturing the complex dynamics of power systems\cite{sofi2022case}. To model this complexity, time series models like Auto-Regressive Moving Average (ARIMA) \cite{pmdarima} various deep learning techniques have been introduced such as  Artificial Neural Networks (ANN) \cite{mcculloch1943logical}, Recurrent Neural Networks (RNN) \cite{Rumelhart_Hinton_Williams_1985}, Long Short-Term Memory (LSTM) \cite{hochreiter1997long}, and Gate Recurrent Units (GRU)\cite{gao2016deep}. The models work to improve the accuracy of load forecasts by leveraging large datasets and discovering hidden patterns to predict future values.

Recently Transformer models \cite{DBLP:journals/corr/VaswaniSPUJGKP17} have revolutionized machine learning due to their unique architecture. Because of the capability to run parallelly across multiple GPUs, they perform more efficiently compared to other deep learning models and take less time to train compared to  sequential models such as LSTMs\cite{Miao_2022}. Furthermore, as the Transformer model generates results after training through backpropagation, they can generate future results using a larger reference window in comparison to RNNs, LSTMs, and GRUs\cite{DBLP:journals/corr/abs-1810-04805}. This window gives Transformers a better ability to identify long-range dependencies in sequences and better resistance towards the vanishing gradient problem \cite{Hochreiter_1998} compared to other deep learning models. The Transformer's strength in identifying long-range dependencies made them the optimal model for natural language processing and they are used in machine translation, text generation, speech recognition, and more. 

Like any other deep learning model, their performance depends on the chosen hyperparameters. In this work we utilized metaheuristics Genetic Algorithm \cite{Holland_1992}, Differential Evolution \cite{Storn_Price_1997}, and Particle Swarm Optimization\cite{488968} to identify ideal hyperparameters. Although hyperparameter search techniques like Grid Search \cite{lavalle2004relationship}, Random Search \cite{bergstra2012random}, and Bayesian Optimization \cite{snoek2012practical} are substantial improvements to manual tuning, they are inferior to the metaheuristics discussed this paper. The metaheuristics are more efficient than grid search and random search and more robust and scalable than Bayesian Optimization. Furthermore, these algorithms can be applied to nonlinear, nonconvex, and noncontinuous functions\cite{Cao_Nguyen_Nguyen_Truong_Nguyen_2023} \cite{2023arXiv230902600S}.

Traditional Transformers take a sequence of tokenized inputs. For Natural Language Processing these inputs are words but can be generalized to other sequential data for other tasks. These tokens are then run through several encoder and decoder layers. Encoders process the input using the self-attention mechanism to find dependencies between tokens and positional encoding to maintain the ordering of tokens. The decoders then generate output token sequences using similar self-attention mechanisms, but also a unique encoder-decoder attention layer that allows it to read the encoded information.
\par
In this work, we created a custom Transformer Neural network model. Our model only uses the encoder of the Transformer and uses it to enhance Deep Learning Models for load forecasting. This research is unique by investigating the Transformer's Attention Mechanism capabilities outside of the usual scope of natural language processing. We identify that the Transformer's abilities in long-range dependencies can be applied to load forecasting.
\par
Our work seeks to fill the void and propose Differential Evolution optimized custom Transformer Neural Networks specifically designed for load forecasting. To evaluate the results we also integrated Particle Swarm Optimization and Genetic Algorithm with the Transformer Neural Networks to benchmark against our proposed Differential Evolution integrated Transformer Neural Network. In particular, our work is the first to propose a Differential Evolution-based hyperparameter tuning scheme for a Transformer-based Neural Network model for load forecasting. 

\section{Preliminaries}
\subsection{Differential Evolution}

Differential Evolution (DE) is a stochastic population-based optimization algorithm developed by Rainer Storn and Kenneth Price in 1997. It is used to find approximate solutions to a wide class of challenging objective functions. DE can be used on functions that are nondifferentiable, non-continuous, non-linear, noisy, flat, multi-dimensional, possess multiple local minima, contain constraints, or are stochastic \cite{5773566}. A general problem formulation that DE could solve is:
\begin{center}
    For objective function $f:X \subseteq \mathbb{R}^n \rightarrow \mathbb{R}$ where $X \neq \emptyset$ find
    $s \in X$ s.t. $f(s) \leq f(x)$ $ \forall x\in X$
    where $f(s) \neq -\infty$ 
\end{center}
Its versatility comes from its unique implementation that does not require the gradient of the function. DE obtains a minimum solution by initializing a set of candidate solutions and iteratively improving each solution by applying various genetic operators \cite{inproceedings}.

\subsubsection{Initialization}
Suppose $f$ has $D$ parameters. An $N$-sized candidate solution population is initialized, with each candidate solution modeled as $x_i$, a $D$-parameter vector.

\begin{center}
$x_{i, G} = [x_{1, i, G}, x_{2, i, G} ... x_{D, i, G}]$ where $i = 1,2...N$ \newline and $G$ is the generation number
\end{center}

Each index $x_{j, i, G}$ with $j = 1, 2...D$ represents a parameter to be manipulated approximate a solution to the objective function \cite{10.5555/1121631}. During the initialization of the first generation, each parameter for all candidate solutions is set randomly from bounds $[x_{j}^{L}, x_{j}^{U}]$.
\begin{center}
$x_{j}^{L} \leq x_{j, i, 1} \leq x_{j}^{U}$
\end{center}

\subsubsection{Mutation}

A mutation is a stochastic change that expands the candidate solution search space. Mutations are used in DE to prevent the algorithm from converging upon a local optimum \cite{article}. In the original mutation scheme devised by Storn, a mutant vector $v_{i}$ is created from randomly sampling three candidate solution vectors $v_{r_{1}}$, $v_{r_{2}}$, $v_{r_{3}}$ such that $r_{1}, r_{2}, r_{3}$ and $i$ are distinct. The mutant vector is obtained by adding the weighted difference of two of the vectors to the third.

\begin{center}
    $v_{i, G+1} =v_{r_{1}, G} + F \times (v_{r_{2}, G} - v_{r_{3}, G})$
\end{center}

\noindent $F \in [0, 2]$ represents the scale factor controlling the magnitude of the mutation.

\subsubsection{Crossover}

Crossover is how successful candidate solutions pass their characteristics to the following generations. A trial vector $u_{i, G+1}$ is created by combining the original vector $x_{i, G}$ and its corresponding mutant vector $v_{i, G+1}$.  A widely used crossover scheme is described below:\cite{Georgioudakis_Plevris_2020}. 
\begin{center}
$u_{j, i, G+1} = 
\left\{
    \begin{array}{lr}
        v_{j,i, G+1}, & \text{if } p_{rand} ~ U(0, 1) \leq CR \\
        x_{j, i, G} & \text{else} 
    \end{array}
\right\}$
\end{center}

\noindent Each $j=1, 2....D$ and $v_{i, G+1} \neq x_{i, G}$

\subsubsection{Selection}
Given both the initial target vector and generated trial vector, the fitness of each is evaluated using the initial objective or cost function $f$. The vector with the lower cost is passed to the next generation.

\begin{center}
$x_{i, G+1} = 
\left\{
    \begin{array}{lr}
        u_{i, G+1}, & \text{if } f(u_{i, G+1}) \leq f(x_{i, G}) \\
        x_{i, G} & \text{else} 
    \end{array}
\right\}$
\end{center}
\vspace{0.25cm}
\par
The Differential Evolution Algorithm is illustrated in Figure 1 below.

\begin{figure}[htbp]
\centering
\includegraphics[height=5.25cm,width=8cm]{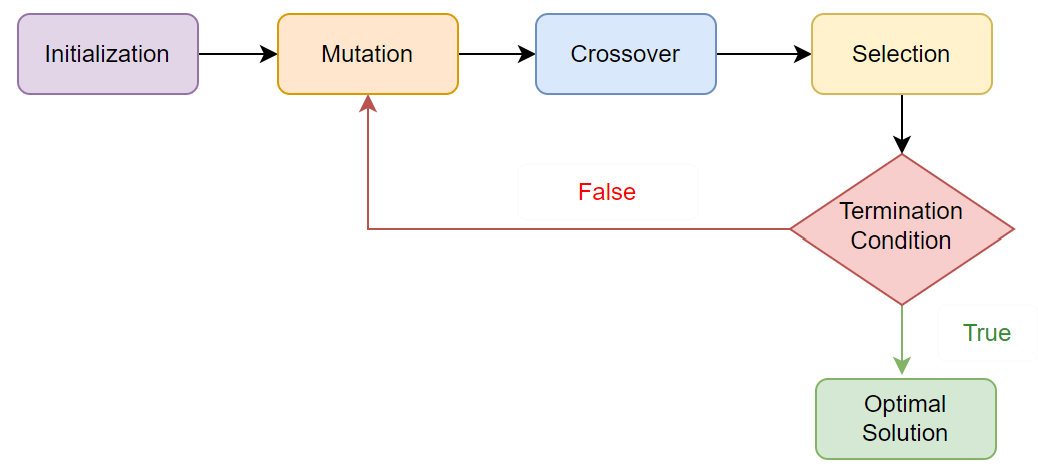}
\caption{Differential Evolution Algorithm}
\label{Fig-1}
\end{figure}
\noindent Mutation, crossover, and selection are cycled until either the maximum number of generations is attained or the candidate solutions meet a predefined accuracy threshold as defined.

\begin{figure*}[t]
\centering
\includegraphics[height=8.15cm,width=16.5cm]{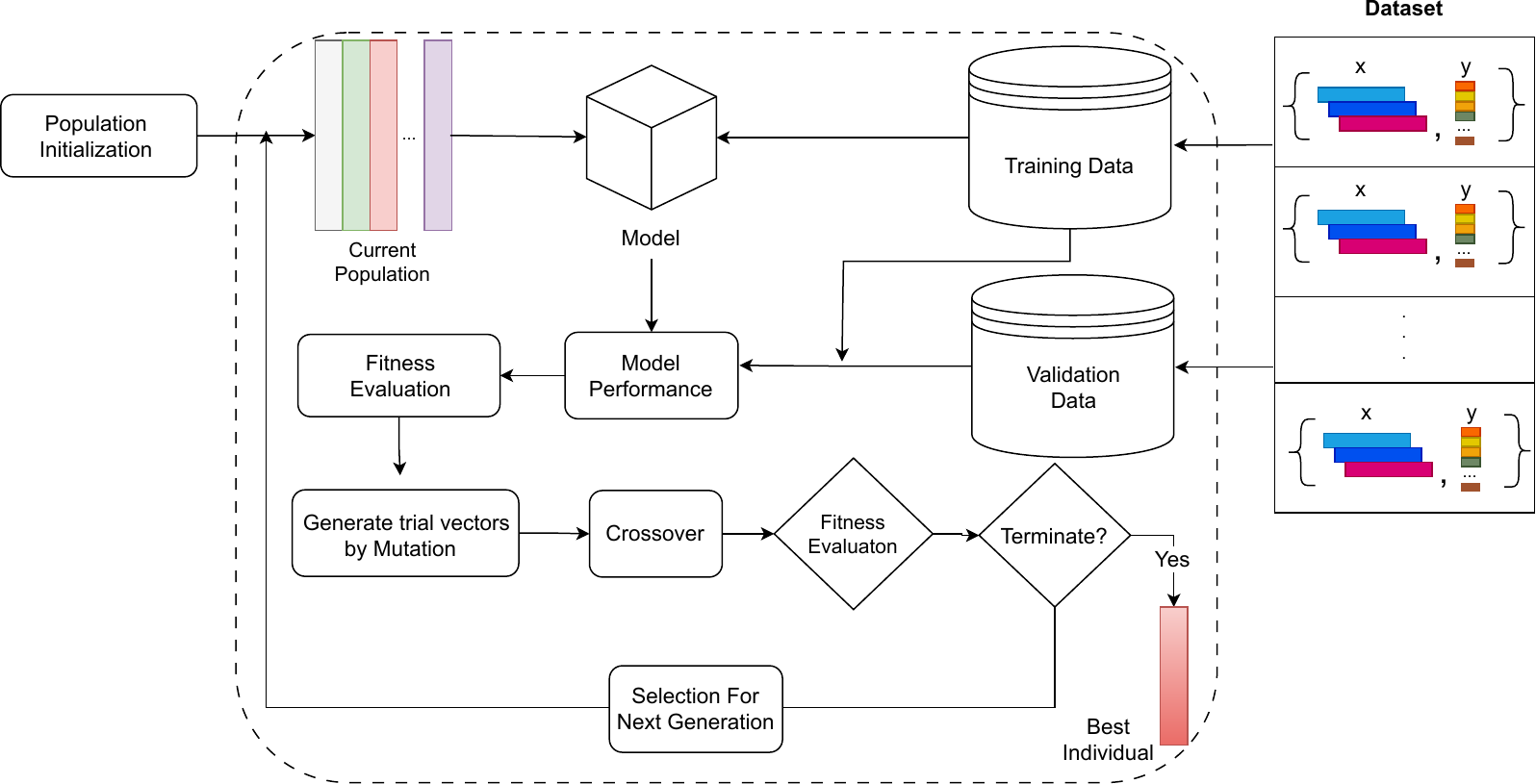}
\caption{Mechanism of the Differential algorithm based hyper-parameters selection approach for the Load Forecasting task.}
\label{Fig-2}
\end{figure*}
\newpage

\section{Proposed Approach}
This section describes the implementation of integrating metaheuristics with the Transformer-based Neural Network for load forecasting. Each metaheuristic is used to identify the optimal set of hyperparameters and the efficacy of the hyperparameters is measured using the Mean Squared Error (MSE) and Mean Average Percentage Error (MAPE) metrics. The integrated Differential Evolution mechanism selection strategy for the hyperparameters is outlined in Figure 2 above.
\subsection{Transformer-based Neural Network}
We implemented the Transformer-based Neural Network by building a sequential model and sequentially adding layers. 
The input layer is initialized with 36 nodes and then passed through a dense time-distributed Transformer-based Neural Network layer of 64 nodes. Next, we have an 8-headed attention layer with a dimension of 64 and a dropout rate of 0.1, where the output of the previous operation is applied. The result is then flattened and run through 2 dense layers containing 64 nodes before returning through a 24-node output layer. The activation function used for all cases, except for the output layer, is Rectified Linear Units (ReLU) \cite{agarap2018deep} (modeled below) except for the output layer.
$$f(x) = max(0, x)$$
The output uses a Linear Activation Function. 
$$f(x) = x$$

All Transformer-based Neural Networks use the implemented form of metaheuristics algorithms to optimize the batch size, learning rate hyperparameters, and epoch. Optimization is done by minimizing loss using Mean Squared Error. The metaheuristic-optimized Transformer-based Neural Networks are assessed by comparing the MAPE for each set of hyperparameters found.

The entire architecture of the proposed model is portrayed in Figure 3 below.

\begin{figure}[htbp]
    \centering \includegraphics[height=6cm,width=6.5cm]{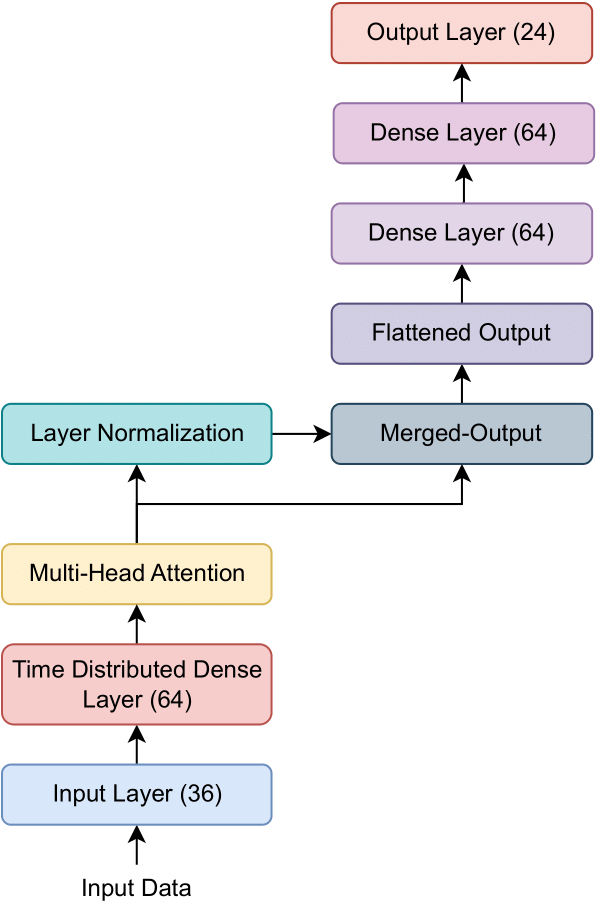}
    \caption{The proposed Transformer-based deep learning model}
    \label{Fig-3}
\end{figure}
\par
The multi-headed attention is primarily used for the model to simultaneously operate different parts of the input sequence, improving the performance. The normalization layer is then passed on top of the attention layer to make the model robust. This avoids the scenario of the model relying too much on specific features, which reduces over-fitting. The best-optimized set of hyperparameters obtained from each metaheuristic algorithm is then applied to each Transformer-based Neural Network and subsequently tested on the test dataset to generate the results. 
\vspace{0.25cm}

Each model's best set of batch size epoch and learning rate are summarized in Table I below:
\begin{table}[htbp]
    \centering
    \caption{Best Set of Hyperparameters for Transformer}
    \label{table:1}
    \begin{tabular}{|c|c|c|c|}
      \hline
      \textbf{Metaheuristics} & \textbf{Batch Size} & \textbf{Epoch} & \textbf{Learning Rate} \\
      \hline
      Genetic Algorithm & 80 & 844 & 0.0001 \\
      \hline
      Particle Swarm & 173 & 35 & 0.3109 \\
      \hline
      Differential Evolution & 24 & 1000 & 0.1 \\
      \hline
   \end{tabular}
\end{table}

\par
After passing the attention layer, the output is flattened. This means it is reshaped from a 3D to a 2D tensor, which allows the subsequent layers to treat the output as a sequence of 2D inputs. The flattened output is then passed through two dense layers, each consisting of 64 layers. These dense layers and increased nodes allow the model to capture more complex patterns and relationships in the data.
\par
The output layer contains 24 nodes, as the model is designed to produce a prediction 24 hours ahead. The custom transformer model architecture developed is intended for the specific task of short-term load forecasting. Its primary aim is to aid the industry by operating on load data to predict variations in various load parameters.
\section{Experimental Details}
\subsection{Dataset Description}
For this project, the \textit{Load Dataset}\footnote{Dataset Link: \url{https://doi.org/10.7910/DVN/O8QA5H}} was curated using meteorological data scraped from the official website of the Government of Canada\cite{Canada_2023}. The dataset covers the period from 1\textsuperscript{st} January 2017 to 4\textsuperscript{th} July 2023 in Ottawa, Ontario. It contains 19 variables, capturing details such as date, time (in 24 hours), year, quarter, month, week of the year, day of the year, state holiday, hour of the day, day of the week, day type, temperature (in $^\circ$C), dew point temperature (in $^\circ$C), relative humidity (\%), wind speed (in km/h), visibility (in km), precipitation amounts (in mm), daily peak (in MW), and hourly demand (in MW). In total, there are 96,432 rows, with each row representing data for a specific hour.

\subsection{Preprocessing}

During the preprocessing stage, we addressed missing data in the compiled dataset. Since the precipitation column had significant missing information, it was excluded from the analysis. Regarding the temperature, only $0.03\%$ of the data was missing. To forecast predictions up to 24 hours into the future, we used 3 hours of past data. The data was standardized using the StandardScaler function from the sklearn.preprocessing library \cite{pedregosa2011scikit}.

The dataset was then split into three subsets: the training dataset, denoted as $D_{train}$, the validation dataset, denoted as $D_{val}$, and the testing dataset, denoted as $D_{test}$. The training dataset covers the period from January 1st, 2017, to December 31st, 2020. Within this dataset, 25\% of the data was allocated to the validation dataset. The remaining data, extending until July 14th, 2023, constitutes the testing dataset.
\subsection{Experimental Setups}
The experiments of this work are implemented in Python
3.10.11 using three libraries : Tensorflow 2.11.0, Tensorflow
built in Keras, and Numpy 1.21.

\section{Results and Discussion}

We obtained the mean absolute percentage error (MAPE) using the proposed approach to implement the differential evolution-based hyperparameter tuning of the Transformer-based deep Neural Network offered in the preceding section. This MAPE was compared to the MAPEs generated from the proposed approach with the genetic algorithm and particle swarm optimization-based hyperparameter tuning of the custom architecture. The codes used in this paper are linked below\footnote{Code Link: \url{https://github.com/AnuvabSen1/Meta-Transformer}}.
\par
The Standard scaler has been used to improve the convergence and stability of seasonal data during model training. This scaler prevents features of larger sizes from dominating the training process and also normalizes the dataset, allowing the model to learn effectively from the data. These steps are
necessary for improving forecasting models.
\par
The mean squared error (MSE) is used to measure the fitness of the differential evolution algorithm. 
\par
MSE serves as the loss function and is plotted against the number of epochs for the entire training duration as shown in Figure 4.
\vspace{0.15cm}

\begin{figure}[htbp]
  \label{Fig-4}
  \centering \includegraphics[width=8cm, height=5.65cm]{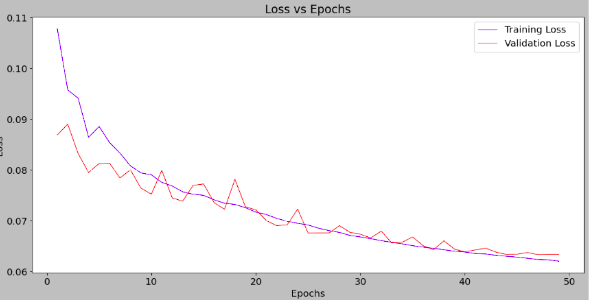}
  \caption{Training \& Validation Loss vs Epochs plots for the Transformer-based Neural Network DE model}
\end{figure}

\begin{figure*}[t]

\centering

\includegraphics[height=9 cm,width=16.5cm]{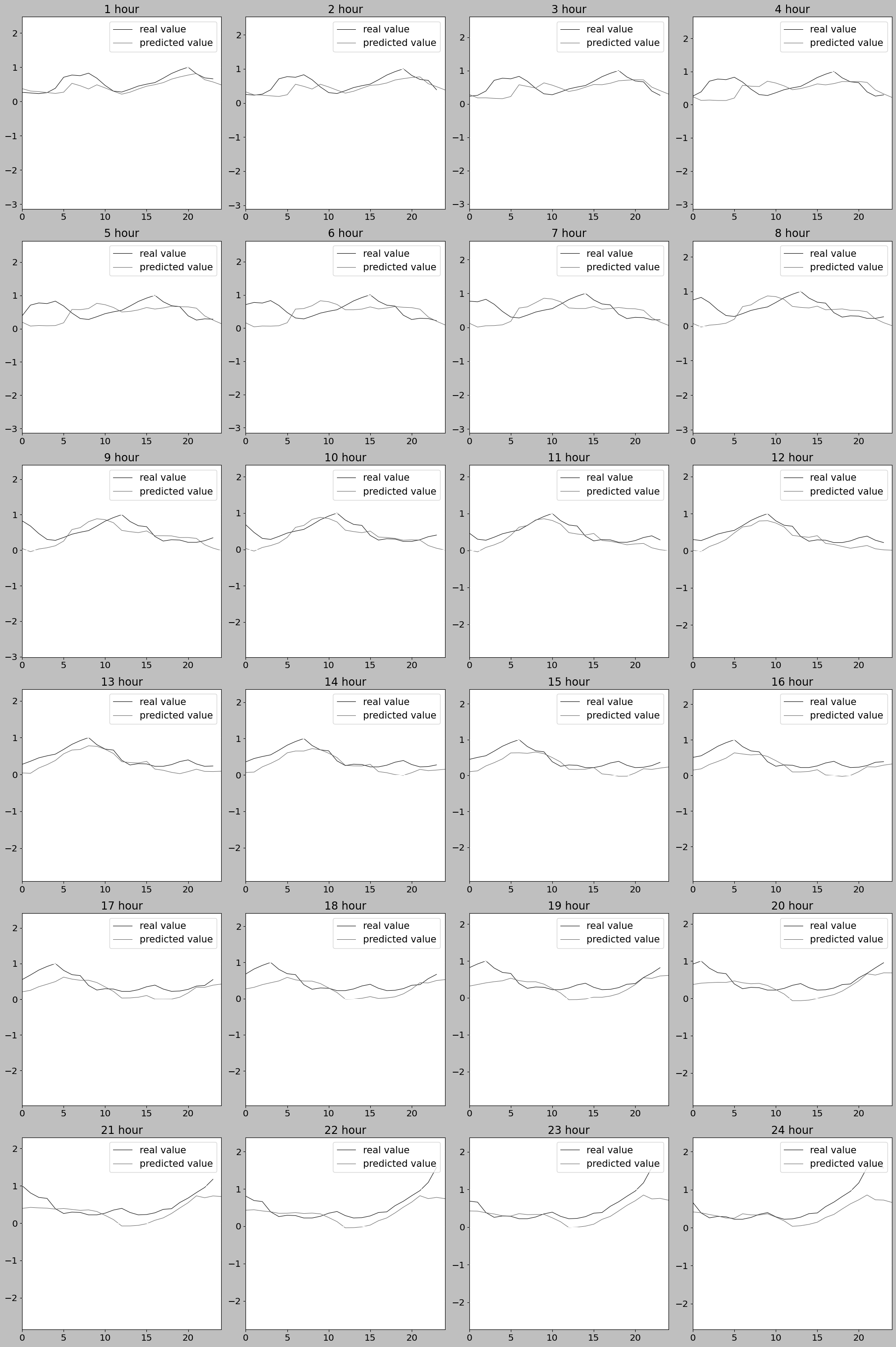}



\caption{Predicted plots for hourly demand for next 24 hours starting from the $N$-th hour for Transformer-based Neural Network DE Model.}

\label{Fig-5}

\end{figure*}
\vspace{0.15cm}
The plot helps us to observe how the loss changes over time and whether the model is optimizing or overfitting.

Mean Absolute Percentage Error (MAPE) is used to gauge the accuracy of the entire model. It provides a measure of the average percentage difference between predicted values and the actual values. 

Table II below provides us with a comparison of MAPE among various metaheuristic optimization algorithms used here.
\begin{table}[htbp]
    \caption{Comparison of Metaheuristics and MAPE for Best Set of Hyperparameters for Transformer}
    \label{table:2}
    \centering
    \begin{tabularx}{0.48\textwidth}{|>{\centering\arraybackslash}X|>{\centering\arraybackslash}X|} 
      \hline
      \textbf{Metaheuristics} & \textbf{MAPE} \\
      \hline
      Manual Selection & 2.07 \\
      \hline
      Genetic Algorithm & 1.31 \\
      \hline
      Particle Swarm & 1.28 \\
      \hline 
      Differential Evolution & 1.11 \\
      \hline
    \end{tabularx}
\end{table}
\par
The results prove that Differential Evolution (DE) algorithm outperforms the Genetic Algorithm (GA) and
Particle Swarm Optimisation (PSO) in terms of mean absolute percentage error (MAPE).
\par
Differential Evolution's superior ability can be attributed to a few factors. DE can more effectively explore the search space and exploit the promising regions for optimal solutions using its various genetic operators, therefore producing most desirable results. To visually understand the results and accuracy of the load forecasting model proposed here we have used two plots. 
\par
The first plot provides us with a 24-hour prediction for the best-performing DE on the Transformer-based Neural Network model as shown in Figure 6. 
\par This shows that DE on Transformer-based Neural Network gives a fairly accurate prediction on Test data.
The second graph plots the hourly demand variation for 24 hours starting from the $N$ \textsuperscript{th} hour shown in Figure 5.
\par
The plots indicate that the accuracy decreases as N increases or as further in time we want to predict the less accurate results we obtain.
The mutation operator sets random disturbances to ensure the prevention of early convergence towards a local minimum.
The crossover operator passes on successful attributes to accelerate the convergence process even further.
\begin{figure}[htbp]
  \label{Fig-6}
  \centering
  \includegraphics[width=8.0cm, height=4.5cm]{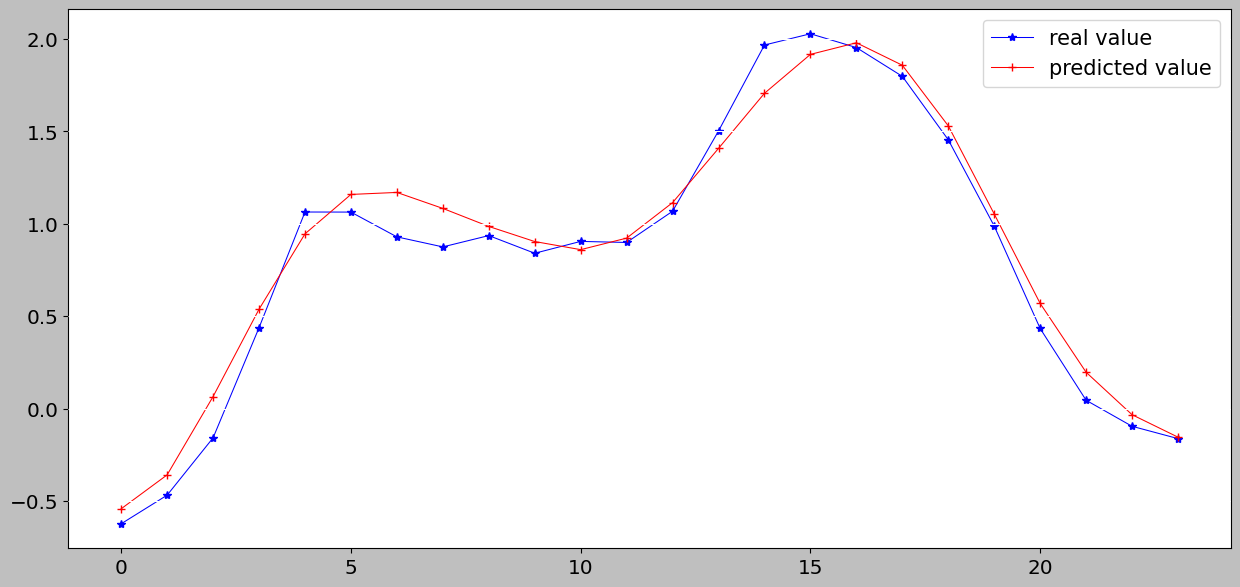}
  \caption{24-hours ahead forecast plot for the Transformer-based Neural Network DE model}
\end{figure}
The selection operator preserves the fittest candidate solutions to improve the quality of results. 
\par
The results ascertain that metaheuristic optimization algorithms consistently outperform the manual selection method of selecting hyperparameters. Particle Swarm Optimization (PSO) performs better than Genetic Algorithm (GA) but falls short behind Differential Evolution (DE). PSO suffers from rapid convergence, limiting its ability to reach the global optimum, which could be an explanation for its performance.
\par

\section{conclusion and future work}
This paper applies several metaheuristic algorithms to a custom Transformer-based Neural Network to find the optimal hyperparameters. This selection method was proven to be far more efficient and accurate than manual selection. Amongst the metaheuristics tested, Differential Evolution proved to be the best because of its mutation and selection operators which not only allowed the algorithm to thoroughly search the sample space but the filter and refine the best solutions. Differential Evolutions performance was then followed by Particle Swarm Optimization and finally Genetic Algorithm.

Due to possessing limited computational resources, each metaheuristic algorithm couldn't be applied to sufficiently large populations over many generations. If this research is extended with more powerful devices, future studies over larger populations and more generations will corroborate our findings. Future study may investigate the performance of other alternative metaheuristic algorithms on hyperparameter tuning for similar deep learning models, across a wide range of forecasting tasks.

\section{acknowledgement}
This work was made possible through the generous backing of Mitacs and the dedicated support of Dr. Chi Tang, Associate Professor at McMaster University. Their invaluable support enabled Anuvab Sen to embark on an enriching journey of undergraduate research at McMaster University, Canada.

\vspace{12pt}
\color{red}

\end{document}